\documentclass{article}

\usepackage[preprint]{neurips_2026}

\usepackage[utf8]{inputenc} 
\usepackage[T1]{fontenc}    
\usepackage{hyperref}       
\usepackage{url}            
\usepackage{booktabs}       
\usepackage{amsfonts}       
\usepackage{nicefrac}       
\usepackage{microtype}      
\usepackage{xcolor}         

\usepackage{multirow}
\usepackage{enumitem}
\usepackage{makecell}
\usepackage{amsmath}
\usepackage{amssymb}
\usepackage{graphicx}
\usepackage{subcaption}
\usepackage{gensymb}
\usepackage{comment}
\usepackage{diagbox} 
\usepackage{booktabs}
\usepackage{wrapfig}

\newcommand{\Tref}[1]{Table~\ref{#1}}

\newcommand{\Fref}[1]{Figure~\ref{#1}}
\newcommand{\sref}[1]{Sec.~\ref{#1}}

\newcommand{\textblock}[1]{\vspace{0pt}\noindent\textbf{#1}\hspace{0.2em}}

\def\eg{\emph{e.g.}}

\def\ourPaperTitle 
{Scenes as Objects, Not Primitives: Instance-Structured 3D Tokenization from Unposed Views}

\title{\ourPaperTitle}

\author{%
Mijin Yoo$^1$\thanks{Equal contribution.} \quad
In Cho$^1$\footnotemark[1] \quad
Subin Jeon$^2$ \quad 
Jiwoo Lee$^1$ \quad 
Eunbyung Park$^1$ \quad 
Seon Joo Kim$^1$\thanks{Corresponding author.}\\[3mm]
$^1$Yonsei University \quad $^2$Seoul National University\\
}

\begin{document}
\maketitle
\begin{figure}[h!]
\centering
\includegraphics[trim={0 0 0 60pt}, width=0.85\linewidth]{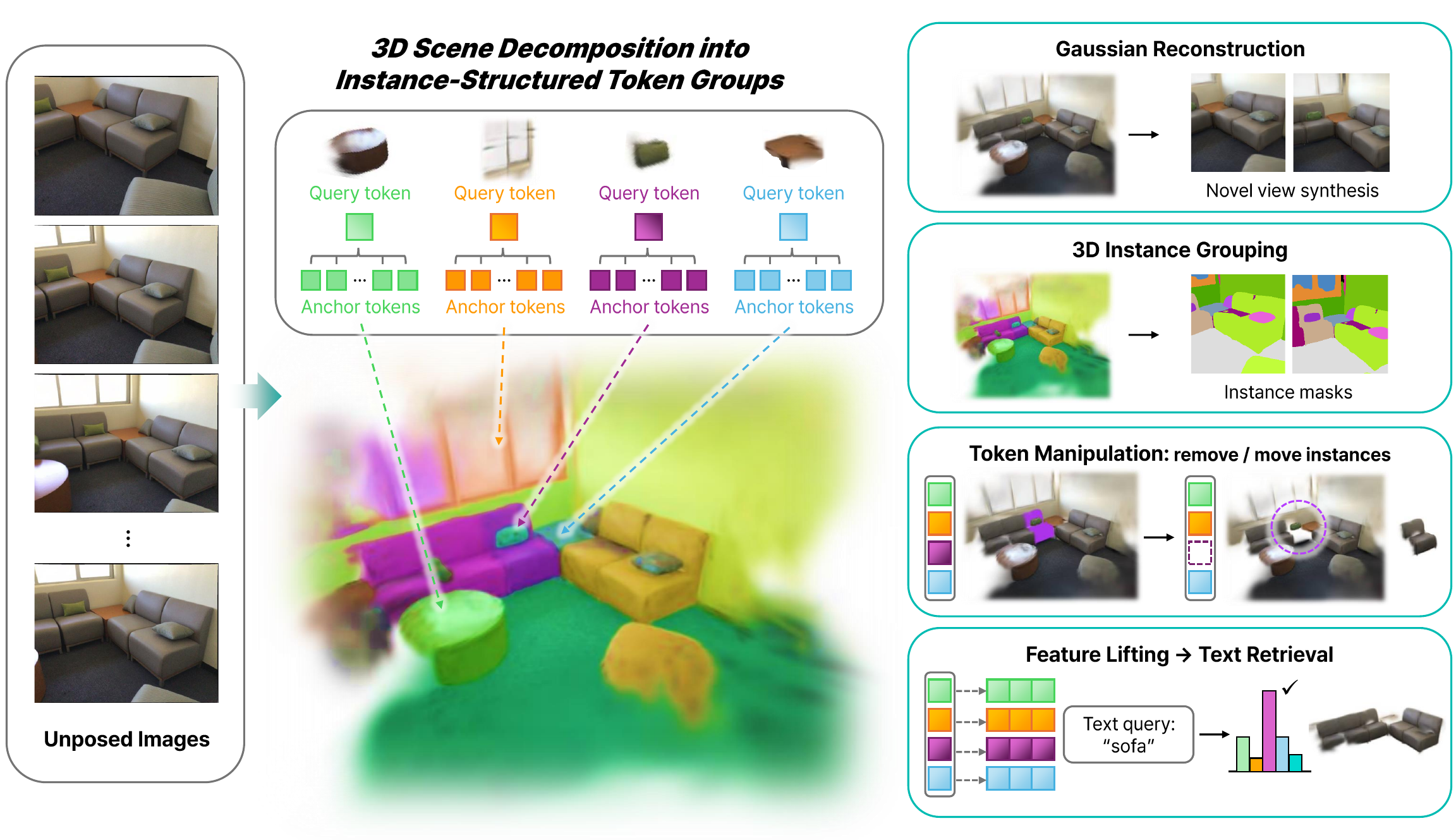}
   \vspace{-7pt}
   \caption{
   Our model maps unposed multi-view images to instance-structured 3D token groups, which make instances a native interface of the 3D representation. The token groups support novel-view synthesis, 3D instance segmentation, instance-level manipulations and open-vocabulary retrieval. 
   }
   \label{fig:teaser}
   \vspace{-2pt}
\end{figure}
\begin{abstract}
A 3D scene is understood through its objects, not the primitives that compose them. Yet feed-forward reconstruction methods output dense, unstructured sets of points or Gaussians, leaving object-level structure to be recovered after the fact. We propose a feed-forward framework that decomposes a scene into instance-structured 3D token groups directly from unposed multi-view images --- compact object-centric units from which reconstruction, segmentation, and manipulation all follow. Each token group pairs an instance token capturing entity-level identity with anchor tokens that encode local geometry and appearance, which are decoded into a set of 3D Gaussians. This two-level factorization decouples object identity from local appearance, making object instances a native interface of the representation rather than a derived product. The token groups are learned through differentiable rendering with joint reconstruction and segmentation supervision, requiring no 3D annotations. Our feed-forward model surpasses per-scene optimization baselines in class-agnostic instance segmentation while remaining competitive in novel view synthesis. Beyond these metrics, the same token groups directly unlock instance-level scene editing --- removing, translating, or inserting objects by operating on their groups --- as well as efficient open-vocabulary 3D instance retrieval, where retrieval complexity scales with the number of instances rather than primitives.
\noindent Project page: \url{https://yoomimi.github.io/instok3d}
\end{abstract}

\section{Introduction}
A 3D scene is not a bag of primitives. It is a composition of objects whose identities and boundaries give the scene its structure. Yet the dominant paradigm for feed-forward 3D reconstruction produces exactly that: dense, unstructured collections of points or Gaussians, with no notion of what belongs together~\cite{dust3r,vggt,pixelsplat,mvsplat,anysplat,c3g}. For a representation to support object-level reasoning, the entities themselves --- not just the primitives that depict them --- must be first-class units of the representation.

Recent feed-forward 3D reconstruction methods have made remarkable progress in predicting detailed geometry from unposed multi-view images~\cite{dust3r,mast3r,vggt,noposplat,splatt3r,anysplat,c3g}, and a natural next step has been to enrich these reconstructions with semantics by attaching feature vectors from 2D foundation models to each primitive. While effective for local annotation, this strategy does not change the unit of representation. Object-level information remains scattered across many primitives~\cite{lseg,lsm,uni3r,c3g}, the same semantic label is stored redundantly for every element of an entity, and any operation defined over objects --- querying, editing, reasoning --- still requires post-hoc grouping or aggregation~\cite{gaussiangrouping,objectgs,openmask3d,trace3d,lbg,ludvig}.

We view this limitation as a representation mismatch rather than a lack of expressive features. A primitive is a local geometric fragment; regardless of what feature is attached to it, it cannot supply the entity-level context needed to interpret the object it belongs to or provide a meaningful interface for interacting with it. For high-level 3D understanding, the representation should make semantic entities first-class units while preserving access to fine-grained details within each entity.

In this paper, we propose to restructure the representation itself around objects. Given unposed multi-view images, our model decomposes a scene into a compact set of instance-structured 3D token groups in a single forward pass (\Fref{fig:teaser}). Each group pairs an instance token --- which summarizes the identity and extent of an object instance --- with a set of anchor tokens that encode local geometry and appearance, each decoding into a set of 3D Gaussians. The instance tokens expose object-level structure directly, while the anchor tokens preserve the fine-grained detail needed for rendering. This two-level factorization separates what belongs together from how each part looks, making object instances an explicit, manipulable interface of the scene representation.

We learn this representation with a joint reconstruction-segmentation framework built on top of a 3D foundation model. Both objectives are supervised through differentiable 2D rendering: RGB images at novel viewpoints supervise reconstruction, while rendered instance masks supervise grouping. No 3D annotations are required --- the 3D instance structure emerges entirely from 2D supervision.

Once the instance structure is learned, it also provides a natural unit for integrating semantics. Rather than lifting high-dimensional features independently to every Gaussian, we distill 2D foundation model features into the token groups: each group stores a shared instance-level embedding, with lightweight anchor-level residuals capturing spatially varying detail. This yields a compact semantic representation that supports text-based retrieval at the entity level while preserving local specificity.

We evaluate on indoor scene benchmarks across reconstruction, feature lifting, and class-agnostic instance segmentation.
Our feed-forward model surpasses per-scene optimized baselines in instance segmentation, while achieving competitive reconstruction quality. Beyond these results, the same token groups naturally support instance-level scene editing in 3D space --- removing, translating, or inserting objects by directly operating on their token groups, as well as open-vocabulary 3D instance retrieval. These results suggest that structuring 3D representations around objects, rather than primitives, opens a more natural interface for both understanding and interacting with 3D scenes.

\section{Related work}

\textblock{Feed-forward 3D Gaussian reconstruction.}
3D Gaussian Splatting~\cite{kerbl3dgs} represents scenes with efficiently
renderable Gaussian primitives, but the original formulation requires
per-scene optimization from posed images. Feed-forward methods remove this
optimization by predicting Gaussians directly from input views:
pixelSplat~\cite{pixelsplat} and MVSplat~\cite{mvsplat} assume calibrated
views and produce pixel-aligned Gaussians, while recent geometry foundation
models~\cite{dust3r,mast3r,vggt} enable pose-free reconstruction from
unposed image collections. Building on these models, NoPoSplat~\cite{noposplat},
Splatt3R~\cite{splatt3r}, and AnySplat~\cite{anysplat} extend Gaussian
prediction to uncalibrated settings, and Uni3R~\cite{uni3r} and C3G~\cite{c3g}
adapt feed-forward Gaussian representations for scene understanding.
Despite these advances, the representation unit remains low-level: pixels,
point-map elements, Gaussian queries, or individual Gaussians. Such units are
well suited for rendering, but not for human-aligned scene understanding,
where semantics are organized around coherent entities. 
Our token groups shift the primary semantic units from these primitives to entities: we introduce anchor tokens that generate local Gaussians and instance tokens that group anchors into instances.

\textblock{Semantics and instances for multi-view 3D scenes.} 
Instance-aware 3DGS methods often attach identity information directly to the
Gaussian representation. Gaussian Grouping~\cite{gaussiangrouping} optimizes
per-Gaussian identity features, while ObjectGS~\cite{objectgs} reduces
alpha-blending ambiguity using one-hot object-ID channels inherited from
object-aware anchors. These methods produce consistent masks, but require
scene-specific reconstruction and labeling.
Other methods attach semantics to an already formed 3D representation by
lifting 2D masks or features~\cite{openmask3d,sam3d,any3dis,trace3d,lbg,ludvig},
or predict feature-augmented Gaussians in a feed-forward manner
~\cite{lsm, uni3r,c3g}. IGGT~\cite{iggt} and PanSt3R~\cite{panst3r} further move
multi-view instance or panoptic segmentation into the network itself.
However, these outputs are still dense primitive-level masks or features;
persistent 3D object handles require lifting, merging, or rendering after
the fact. In contrast, our token groups are the scene representation itself: anchor
tokens decode local Gaussians, instance tokens bind anchors into instance-level
units, and the resulting groups provide shared handles for feature lifting and
manipulation.

\textblock{Object-centric 3D scene reconstruction.}
Object-centric representation learning models scenes as coherent entities
rather than independent local features, as reflected in slot-based and
set-prediction models~\cite{slotattention,detr,mask2former,kmaxdeeplab}.
In 3D, SlotLifter~\cite{slotlifter} learns object-centric radiance fields
through slot-guided decomposition, while GOCL~\cite{gocl} introduces
object-centric supervision for per-scene optimized 3DGS using a
scene-agnostic object codebook. These works highlight the value of
object-level 3D structure, but the object variables are induced within
radiance fields or optimized Gaussian scenes. In contrast, we learn
instance-structured token groups directly from unposed multi-view images:
anchor tokens generate local Gaussians, and supervised instance tokens group
anchors into human-aligned object-scale units that share semantic features
and serve as entity-level interfaces.
\section{Method}
\label{sec:method}
\begin{figure}[t]
\centering
\includegraphics[width=1\linewidth]{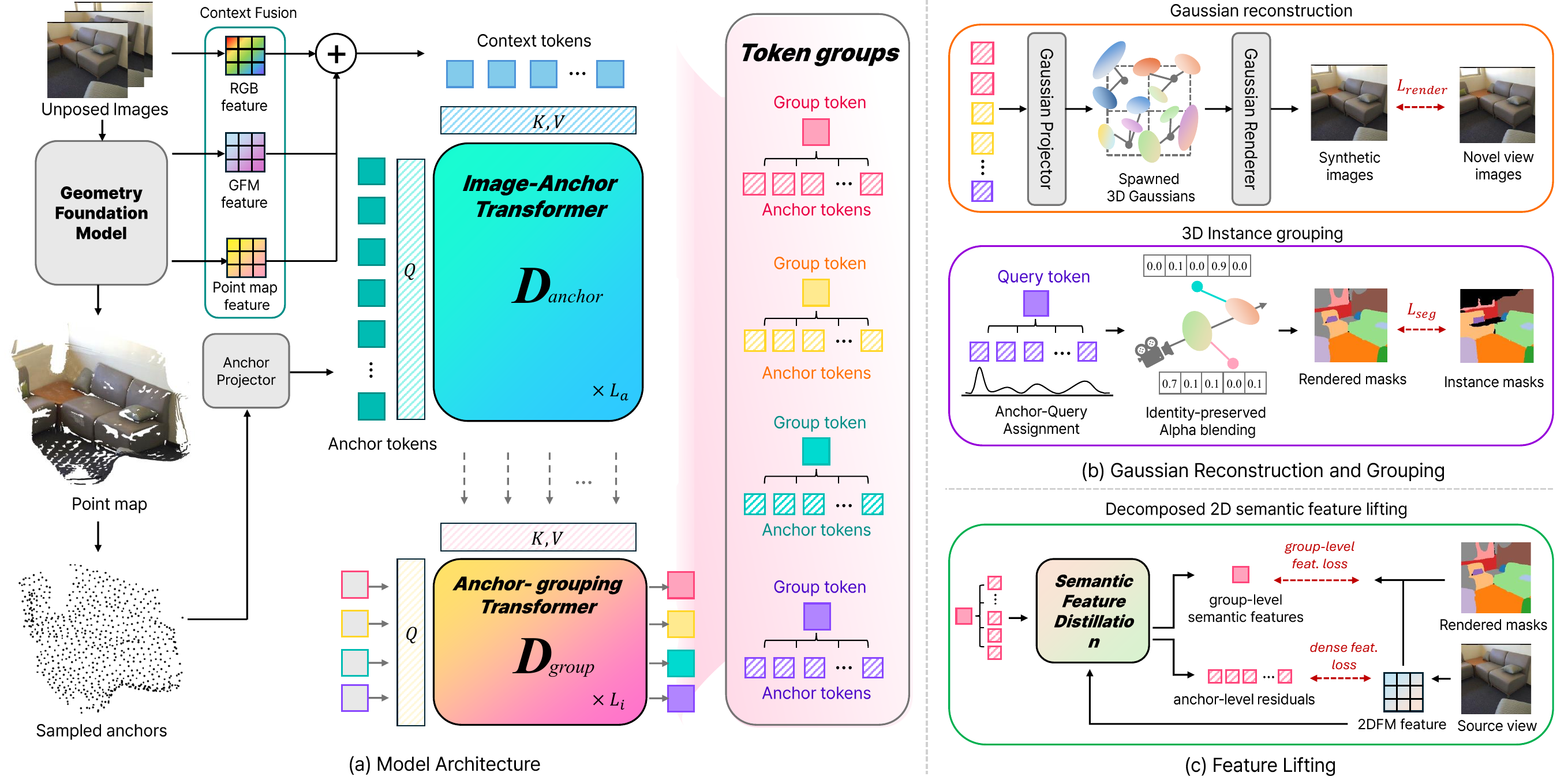}
\vspace{-13pt}
\caption{
\textbf{Overview of the 3D token group framework.} 
\textbf{(a)} Multi-view features and pointmaps from a 3D foundation model are fused into context tokens. The image-anchor decoder $\mathcal{D}_{\mathrm{anchor}}$ decodes anchor tokens from them, and the anchor-grouping decoder $\mathcal{D}_{\mathrm{group}}$ produces group tokens defining instance-level assignments. 
\textbf{(b)} The framework is trained by 2D reconstruction-segmentation supervision: RGB images for anchor tokens, and instance masks for token grouping.
\textbf{(c)} The token groups support semantic feature lifting decomposed into group- and anchor-level components.
}
\label{fig:method}
\vspace{-10pt}
\end{figure}

We present instance-structured token groups, a feed-forward 3D scene representation built from unposed multi-view images. Each group pairs an instance token capturing entity-level identity with anchor tokens that encode local geometry and appearance, which are decoded into a set of 3D Gaussians. \Fref{fig:method}(a) shows the model architecture: a frozen geometry foundation model extracts multi-view features and pointmaps, fused into context tokens. An image-anchor transformer ($\mathcal{D}_{\mathrm{anchor}}$) cross-attends to these context tokens to produce anchor tokens, and an anchor-grouping transformer ($\mathcal{D}_{\mathrm{group}}$) cross-attends to the anchor tokens to produce group tokens, which compete for anchor ownership via softmax assignment (\sref{sec:method:transformer}). \Fref{fig:method}(b) shows the training supervision: RGB rendering losses shape the anchor tokens while instance mask losses shape the group tokens (\sref{sec:method:training}). Figure~\ref{fig:method}(c) shows how 2D foundation model features are distilled into the token groups, decomposed into a shared group-level embedding and low-dimensional anchor-level residuals (\sref{sec:method:distill}).

\subsection{Instance-structured 3D tokenization}
We tokenize unposed multi-view images into instance-structured 3D token groups: a set of anchor tokens that decode into 3D Gaussians, each assigned to one of the learned group tokens.
 
\textblock{Multi-view feature encoding.}
Given $V$ unposed RGB images $\mathcal{I}=\{I_i\}_{i=1}^{V}$, a frozen 3D foundation model~\cite{vggt} extracts multi-view features $F_i \in \mathbb{R}^{H' \times W' \times C}$ and pointmaps $P_i\in\mathbb{R}^{H\times W \times 3}$.
We downsample each pointmap with stride $p$ to obtain patch-aligned 3D coordinates $\tilde{P}_i \in \mathbb{R}^{H' \times W' \times 3}$.
To provide additional appearance and geometry cues, we linearly patchify the RGB images and pointmaps and add them to the foundation model features.
The resulting features are flattened across all views and patches into the context features $X = \{x_j\}_{j=1}^{VH'W'}$, where each $x_j \in \mathbb{R}^C$ is the fused feature at one patch of one view. $X$ serves as the multi-view context for the token group decoder.
 
\label{sec:method:transformer}
\textblock{Token group initialization.}
Anchor tokens are initialized from the patch-aligned 3D coordinates $\tilde{P}$ and their corresponding context features.
We apply farthest point sampling over all patch coordinates to select $K$ anchor positions $\{a_k\}_{k=1}^{K}$.
The $k$-th anchor token at position $a_k$ is initialized as
\begin{equation}
    A_k^{(0)} = x_{a_k} + \phi_{\mathrm{pos}}(a_k),
\end{equation}
where $x_{a_k} \in \mathbb{R}^C$ is the context feature in $X$ at the patch selected as anchor position $a_k$, and $\phi_{\text{pos}}$ is a 2-layer MLP projecting the 3D coordinate $a_k$ to the feature dimension.
We initialize $L$ group tokens $G^{(0)}=\{G_{\ell}^{(0)}\}_{\ell=1}^{L}$ as learnable embeddings.
 
\textblock{Token group decoding.}
The token group decoder consists of two cross-attention transformers: an image-anchor decoder $\mathcal{D}_{\mathrm{anchor}}$ and an anchor-group decoder $\mathcal{D}_{\mathrm{group}}$.
The image-anchor decoder grounds the anchor tokens in the multi-view context by cross-attending to the context features $X$ defined above.  
The anchor-group decoder then updates the group tokens by cross-attending to the decoded anchors, allowing each group token to aggregate object-level information:
\begin{equation}
    A = \mathcal{D}_{\mathrm{anchor}}(A^{(0)}, X),
    \qquad
    G = \mathcal{D}_{\mathrm{group}}(G^{(0)}, A),
\end{equation}
where $\mathcal{D}(Q, Z)$ denotes a cross-attention transformer that updates queries $Q$ using context $Z$.
The decoded anchor tokens are used to reconstruct Gaussian primitives, while the decoded group tokens serve as grouping queries for assigning anchors to object instances.
 
\textblock{Anchor-to-group assignment.}
Each decoded anchor's assignment probability over the $L$ groups is computed by a softmax over dot-product similarities with the group tokens:
\begin{equation}
    \pi_{k,\ell} = \mathrm{softmax}\big(\{\langle A_k, G_{\ell'} \rangle\}_{\ell'=1}^{L}\big)_\ell.
\end{equation}

The softmax induces competition among group tokens for anchor ownership, encouraging each anchor to belong to a single group --- analogous to the slot competition in \cite{slotattention}.
 
\textblock{Gaussian reconstruction.}
Each decoded anchor $A_k$ at position $a_k$ is mapped to $N_g$ 3D Gaussians by a 2-layer MLP that predicts Gaussian attributes: position offsets relative to $a_k$, scale, rotation, opacity, and spherical harmonics.
Each generated Gaussian inherits the assignment score $\pi$ of its parent anchor, yielding $L$ instance-level groups that can be independently rendered and manipulated.
 
\subsection{Training via joint reconstruction and grouping supervision}
\label{sec:method:training}
The decoders are trained entirely through 2D supervision: RGB images supervise reconstruction quality via the anchor tokens, while instance masks supervise grouping via the group tokens.
 
\textblock{Rendering supervision.}
The predicted Gaussians are rendered at target viewpoints and supervised against the ground-truth images with a combined MSE and perceptual loss:
\begin{equation}
    \mathcal{L}_{\mathrm{render}} = \mathcal{L}_{\mathrm{mse}} + \lambda_{\mathrm{lpips}}\,\mathcal{L}_{\mathrm{lpips}}.
\end{equation}
 
\textblock{Grouping supervision via 2D instance segmentation.}
We cast anchor-to-group assignment as a 2D instance segmentation problem.
Each Gaussian inherits the assignment probability $\pi_{k,\ell}$ of its parent anchor; rendering these probabilities through alpha compositing produces $L$ instance probability maps $\{M_\ell\}_{\ell=1}^{L}$. Following standard 2D segmentation pipelines~\cite{detr, mask2former}, we perform Hungarian matching between $\{M_\ell\}$ and the ground-truth 2D instance masks $\{\hat{M}_n\}_{n=1}^{N}$.
Specifically, the matching cost between group $\ell$ and GT instance $n$ is computed on the (detached) rendered probability maps as the same Dice-and-BCE terms used in the mask loss, $\mathcal{C}(\ell,n)=\lambda_{\text{dice}}\bigl(1-\text{Dice}(M_\ell,\hat{M}_n)\bigr)+\lambda_{\text{bce}}\,\text{BCE}(M_\ell,\hat{M}_n)$, and the optimal assignment is obtained by the Hungarian algorithm.
For each matched pair $(\ell, n)$, we apply a per-pixel binary cross-entropy (BCE) loss $\mathcal{L}_{\mathrm{bce}}$ and a Dice loss~\cite{milletari2016vnet} $\mathcal{L}_{\mathrm{dice}}$ between the predicted mask $M_\ell$ and the matched ground-truth mask $\hat{M}_n$:
\begin{equation}
    \mathcal{L}_{\mathrm{seg}} = \lambda_{\mathrm{bce}}\,\mathcal{L}_{\mathrm{bce}} + \lambda_{\mathrm{dice}}\,\mathcal{L}_{\mathrm{dice}}.
\end{equation}
The anchor-group assignment is computed via a softmax over the $L$ groups and an additional zero-valued void channel to account for anchors in non-instance regions.
The softmax with the void channel implicitly discourages unmatched groups from acquiring anchors.
This 2D segmentation objective drives the group tokens to represent coherent entities while organizing anchors into instance-level groups.
The full training loss combines both objectives:
\begin{equation}
    \mathcal{L} = \mathcal{L}_{\mathrm{render}} + \lambda_{\mathrm{seg}}\,\mathcal{L}_{\mathrm{seg}}.
\end{equation}
As the reconstruction is not yet stable early in training, we apply a linear warm-up to $\lambda_{\mathrm{seg}}$ over the first few steps, ensuring the grouping supervision takes full effect once initial geometry has emerged.

\subsection{Decomposed semantic feature distillation}
\label{sec:method:distill}
The instance structure learned in Sections~\ref{sec:method:transformer} and~\ref{sec:method:training} provides a natural basis for integrating semantics. Rather than attaching a high-dimensional feature vector independently to every Gaussian --- as prior methods do --- our token groups enable a decomposed representation: a shared group-level embedding $s_\ell \in \mathbb{R}^{D}$ captures the dominant semantics of each instance, while a low-dimensional anchor-level residual $r_k \in \mathbb{R}^{d}$ ($d \ll D$) accounts for spatially varying detail within the group. This decomposition reduces semantic storage by orders of magnitude while preserving local specificity.

\textblock{Semantic token encoding.}
Given per-view 2D foundation features $\{\Phi_i\}_{i=1}^{V}$, we reuse the trained image-anchor cross-attention $\mathcal{D}_{\mathrm{anchor}}$ to aggregate them into anchor semantic tokens, following \cite{c3g}. The group-level semantic tokens $\{s_\ell\}_{\ell=1}^{L}$ are then produced by an additional anchor-to-group cross-attention transformer whose queries are initialized from the learned group tokens $G_\ell$ via a linear projection. Intuitively, each group token queries the anchor semantic tokens to summarize the semantic content of its assigned instance. The low-dimensional projections of the anchor semantic tokens form the anchor-level residuals $\{r_k\}$, capturing local semantic variations within each group.

\textblock{Group feature alignment.}
At rendering time, each anchor is hard-assigned to its highest-probability group, and the tuple $[\mathrm{onehot}(\pi_{k,\ell}), r_k]$ is attached to every Gaussian spawned from $A_k$. Rendering produces per-view group assignment maps $\hat{S}_{v,\ell}$ and residual maps $\hat{R}_v$. The full per-pixel semantic feature is reconstructed as
\begin{equation}
    F_v(u) = \sum_{\ell} \hat{S}_{v,\ell}(u)\, s_\ell + W_r\, \hat{R}_v(u),
\end{equation}
where $W_r$ projects the residual back to the foundation feature dimension. We optimize two complementary losses:
\begin{equation}
    \mathcal{L}_{\mathrm{distill}}
    = \sum_{v}\sum_{u} \big(1 - \cos\big(F_v(u),\, \Phi_v(u)\big)\big)
    + \sum_{v,\ell} \big(1 - \cos\big(s_\ell,\, \mathrm{avg}_{M^v_\ell}(\Phi_v)\big)\big),
\end{equation}
where $\mathrm{avg}_{\hat{M}^v_\ell}(\Phi_v)$ denotes the average of the foundation features over the rendered mask prediction $M^v_\ell$ of group $\ell$ in view $v$. The first term drives the full reconstructed feature $F_v(u)$ to match the foundation model output at every pixel. The second term is a group-level alignment loss that directly supervises each $s_\ell$ to capture an object-level semantic summary, ensuring that the anchor residuals $r_k$ need only model sub-instance variation rather than carrying the full semantic feature. Together, the two terms enforce a clean division of roles between the group and anchor levels.

\section{Experiments}
We evaluate the proposed token group representations and the tokenization framework on the ScanNet dataset~\cite{dai2017scannet}.
We first assess the tokenizer in terms of reconstruction quality, feature lifting, and class-agnostic instance segmentation.
We then demonstrate the broader applicability of our token groups through the instance-level token manipulations and text-based 3D instance retrieval.

\textblock{Implementation details.}
We set $d=8$ for low-dimensional anchor-level residuals, maximum number of groups $L=100$, $N_g=32$, and $K=1,024$ anchor tokens.
The image-anchor decoder consists of 6 transformer layers, and the anchor-group decoder of 4.
We build our framework upon the pretrained VGGT~\cite{vggt} and follow the training protocol of Uni3R~\cite{uni3r}.
To resolve the scale misalignment between the pretrained VGGT and the ground-truth data, we first fine-tune VGGT with an auxiliary pixel-aligned Gaussian head for a few epochs, which takes less than 6 hours on 4 H200 GPUs.
We then train our tokenizer with two input configurations:
the 2-view setup is trained for approximately 20 hours on 4 RTX A6000 GPUs,
while the 8-view setup is trained for approximately 12 hours on 4 H200 GPUs.
We set $\lambda_{\mathrm{lpips}}=0.05$ and $\lambda_{\mathrm{seg}} = 0.1$.
We apply $\lambda_{\mathrm{seg}}$ warm-up for first 1,500 steps.
Please refer to the supplements for additional implementation details.

\begin{table*}[t]
\centering
\caption{
\textbf{Quantitative reconstruction and feature lifting results} with 2 context views.
\#Sem. units reports the number of primary semantic units (per Gaussian for the baselines, per instance group for our model). Feat. size reports the total number of stored feature scalars.
$^\dagger$: Uni3R stores a compressed 64-dim feature per Gaussian.
$^\ddagger$: Our model stores 512-dim group features with 8-dim anchor residuals.
\textbf{Bold} indicates the best result and \underline{underline} indicates the second-best.
}
\label{tab:recon_feature}
\footnotesize
\setlength{\tabcolsep}{3pt}
\begin{tabular}{l | cc | cc | ccc | ccc}
\toprule
\multirow{2}{*}{Method}
& \multicolumn{2}{c|}{Source view feature}
& \multicolumn{2}{c|}{Target view feature}
& \multicolumn{3}{c|}{Target view reconstruction}
& \multicolumn{3}{c}{Representation cost} \\
& mIoU$\uparrow$ & Acc.$\uparrow$
& mIoU$\uparrow$ & Acc.$\uparrow$
& PSNR$\uparrow$ & SSIM$\uparrow$ & LPIPS$\downarrow$
& \#Gauss
& \#Sem. units
& Feat. size \\
\midrule
LSeg~\cite{lseg}
& 0.470 & 0.789
& 0.482 & 0.793
& -- & -- & --
& -- & -- & -- \\
\midrule
LSM~\cite{lsm}
& 0.527 & \underline{0.810}
& 0.512 & \underline{0.795}
& 24.24 & \underline{0.821} & \underline{0.222}
& 131,072 & 131,072 & 67.1\,M \\
Uni3R~\cite{uni3r}
& 0.540 & \textbf{0.826}
& \underline{0.558} & \textbf{0.827}
& \textbf{25.53} & \textbf{0.873} & \textbf{0.138}
& 131,072 & 131,072 & 8.4\,M$^\dagger$ \\
C3G~\cite{c3g}
& \underline{0.542} & 0.803
& 0.513 & 0.783
& 23.89 & 0.770 & 0.285
& \underline{2,048} & \underline{2,048} & \underline{1.0\,M} \\
\midrule
Ours
& \textbf{0.661} & 0.786
& \textbf{0.657} & 0.789
& \underline{25.28} & 0.771 & 0.238
& \underline{32{,}768} & \textbf{<100} & \textbf{59.4\,K}$^\ddagger$ \\
\bottomrule
\end{tabular}
\vspace{-10pt}
\end{table*}
\begin{figure}[]
\centering
\includegraphics[width=1\linewidth]{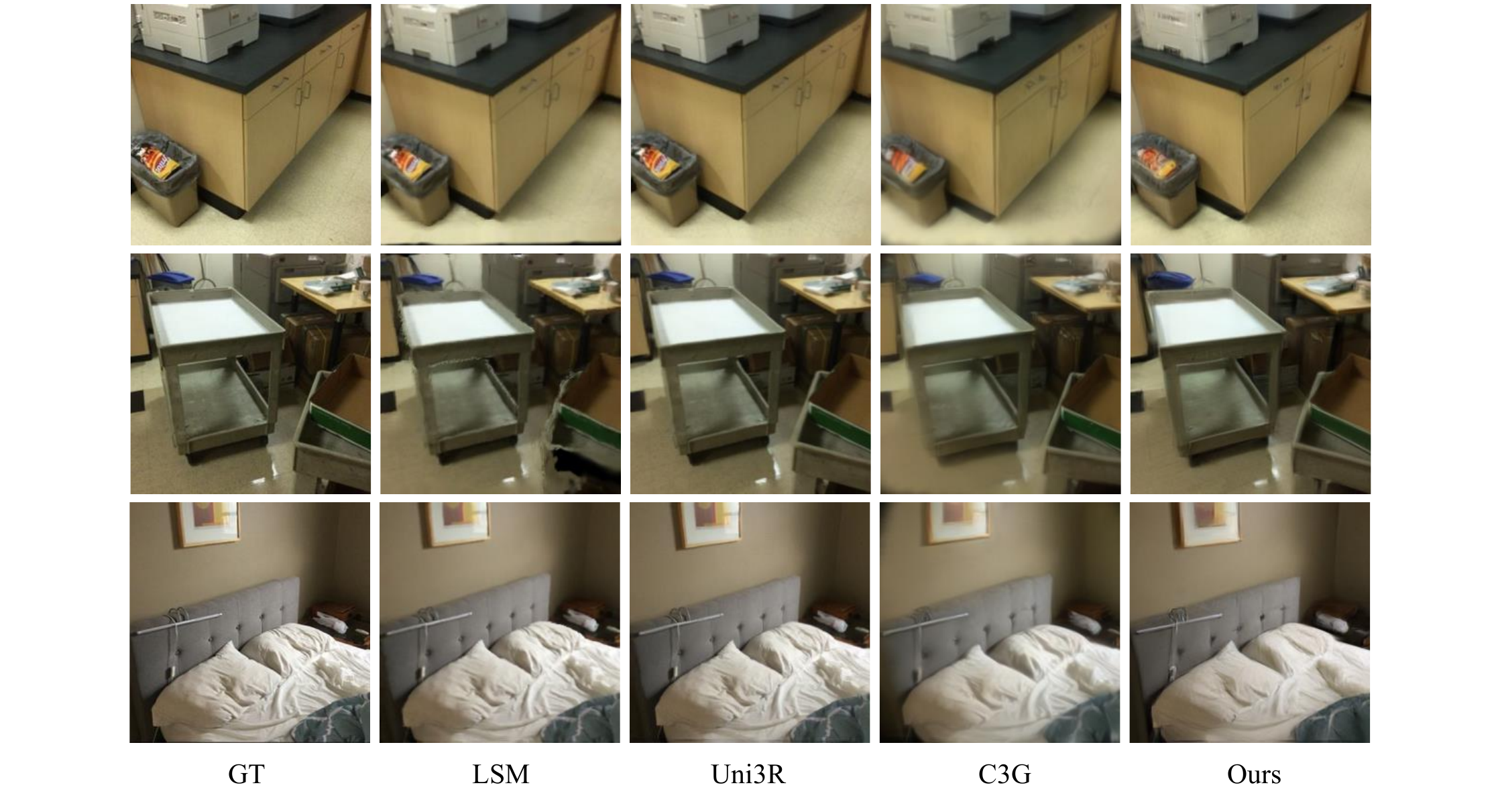}
    \vspace{-20pt}
   \caption{\textbf{Qualitative reconstruction results} with 2 context views.}
   \label{fig:recon}
   \vspace{-10pt}
\end{figure}
\subsection{Tokenization performance}
We evaluate our tokenization framework on three tasks: feed-forward novel-view reconstruction, open-vocabulary feature lifting, and class-agnostic instance segmentation. For reconstruction and feature lifting, we follow the evaluation protocol and source/target camera sampling of LSM~\cite{lsm}, reporting PSNR, SSIM, and LPIPS for reconstruction, and mIoU and pixel accuracy (Acc.) using LSeg~\cite{lseg} for feature lifting, having $D=512$ semantic features. For instance segmentation, we report target-view AP, AP50, and AP25 alongside reconstruction metrics on the same views. We also qualitatively evaluate token-level manipulation and open-vocabulary 3D instance retrieval.

\begin{figure}[]
\centering
\includegraphics[width=1\linewidth]{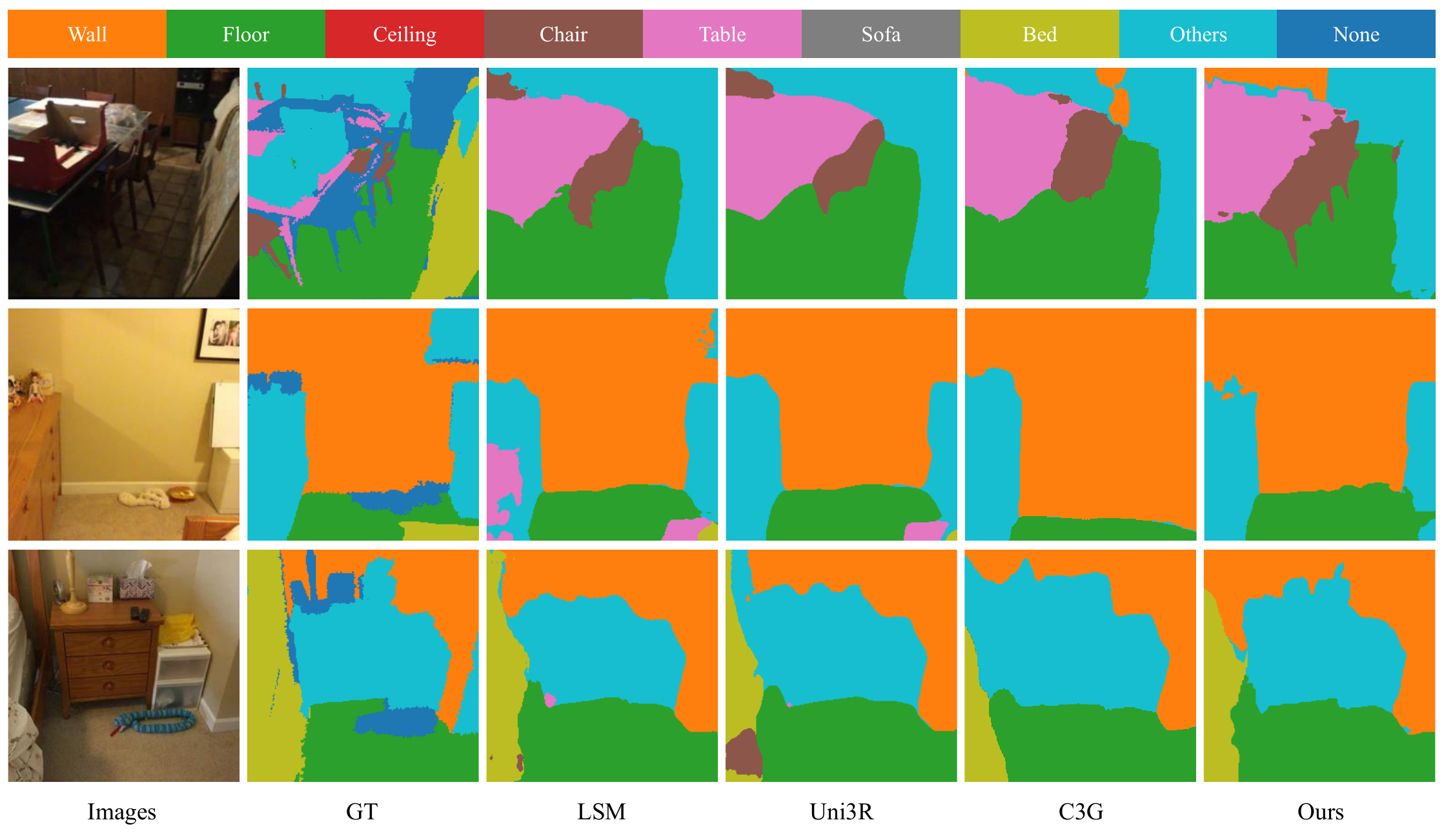}
    \vspace{-20pt}
   \caption{
   \textbf{Qualitative open-vocabulary novel view segmentation results} with LSeg features.
   }
   \label{fig:sem}
   \vspace{-5pt}
\end{figure}
\textblock{Reconstruction and feature lifting results.}
\Tref{tab:recon_feature} reports results on ScanNet with 2 context views. Our model achieves the best feature-lifting mIoU on both source and target views by a clear margin. This advantage reflects a fundamental property of our representation: rather than storing semantic features independently at every Gaussian, our token groups concentrate semantics at the instance level, reducing semantic storage from 8.4M scalars (Uni3R) to 59.4K. On reconstruction, Uni3R and LSM achieve stronger PSNR and SSIM at the cost of 131,072 unstructured per-pixel Gaussians, a scale our compact token-based representation does not match by design.
Importantly, this reconstruction gap narrows considerably in zero-shot transfer to MipNeRF360 (Appendix~\ref{sec:appendix:mipnerf360}), suggesting that the anchor-group structure learns a more transferable scene prior.
Qualitative reconstruction results are shown in Figure 3. Figure 4 shows open-vocabulary feature lifting results, where our token groups produce semantic maps with more coherent object boundaries than competing methods.

\begin{figure}[t]
\centering
\includegraphics[width=1\linewidth]{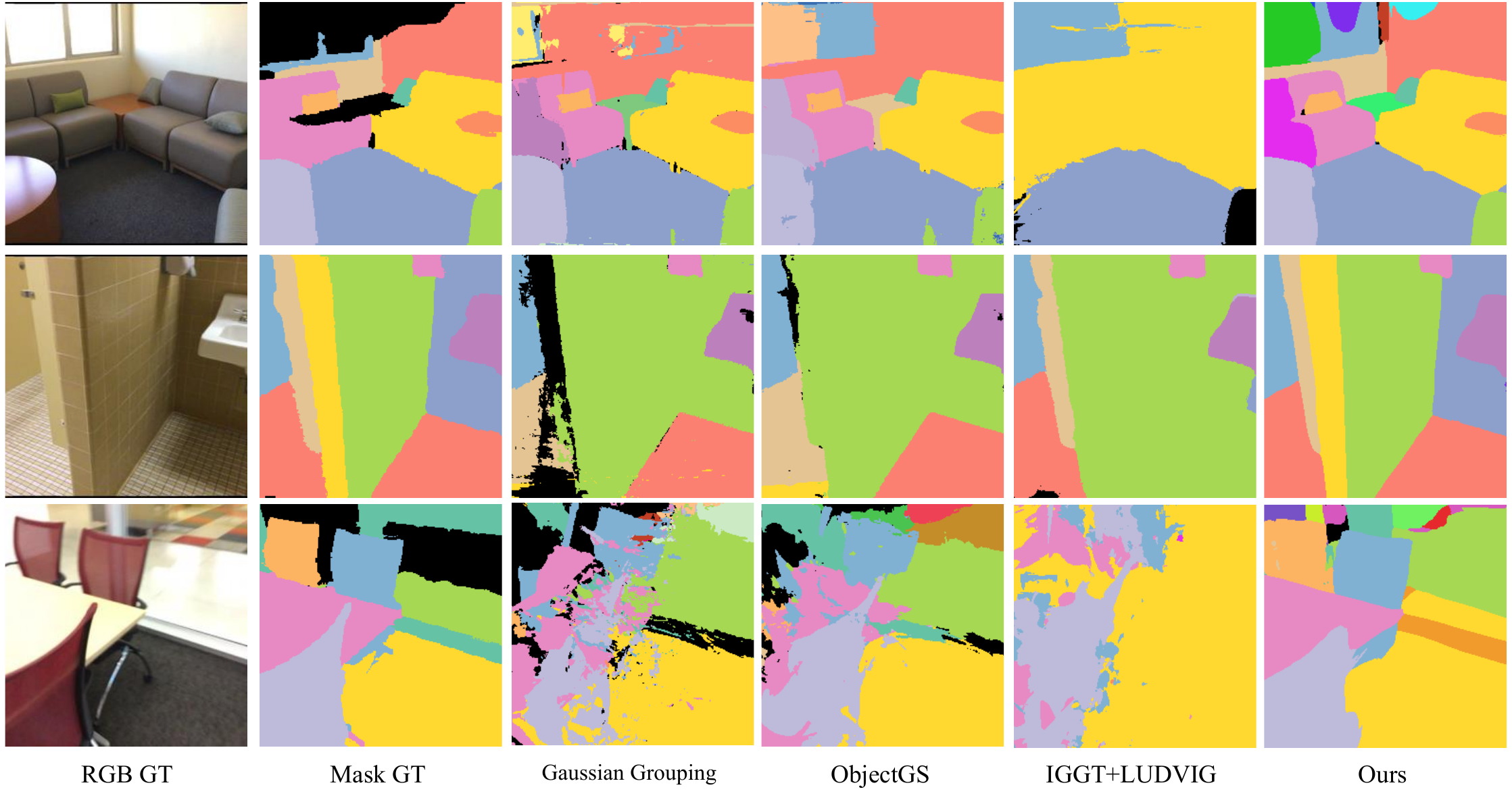}
    \vspace{-18pt}
   \caption{
   \textbf{Qualitative class-agnostic instance segmentation results} with 8 context views.
   }
   \label{fig:class_agnostic_seg}
   \vspace{-5pt}
\end{figure}

\textblock{Class-agnostic instance segmentation results.}
\Tref{tab:instance_seg} reports results with 8 context views. Our feed-forward model achieves the best segmentation across all AP metrics, surpassing both per-scene optimization baselines (Gaussian Grouping~\cite{gaussiangrouping}, ObjectGS~\cite{objectgs}) and the feed-forward method with post-hoc optimization (IGGT [15] + LUDVIG [18]). That a fully feed-forward model trained only on 2D supervision outperforms methods that optimize per-scene suggests that native instance structure is a more effective inductive bias than post-hoc grouping. The qualitative comparisons in \Fref{fig:class_agnostic_seg} make this concrete — competing methods produce fragmented, noisy boundaries particularly on large surfaces, while our token groups yield clean, consistent instance decompositions. Reconstruction quality remains competitive given that all baselines rely on per-scene optimization or pixel-aligned Gaussians, while our model operates in a single forward pass.

\begin{table*}[t]
\centering
\caption{
\textbf{Class-agnostic novel-view instance segmentation and reconstruction} with 8 context views.
AP metrics evaluate target-view instance masks. Reconstruction metrics are
reported on the same target views. \textbf{Bold} indicates the best result and \underline{underline} indicates the second-best.
}
\label{tab:instance_seg}
\footnotesize
\renewcommand{\arraystretch}{1.15}
\setlength{\tabcolsep}{4pt}
\begin{tabular}{l|l ccc | ccc}
\toprule
\multirow{2}{*}{Type}
& \multirow{2}{*}{Method}
& \multicolumn{3}{c|}{Instance Segmentation}
& \multicolumn{3}{c}{Reconstruction} \\
& & AP$\uparrow$ & AP$_{50}\uparrow$ & AP$_{25}\uparrow$
& PSNR$\uparrow$ & SSIM$\uparrow$ & LPIPS$\downarrow$ \\
\midrule
\multirow{2}{*}{\textit{Per-scene optimization}} &
Gaussian Grouping~\cite{gaussiangrouping}
& 0.139 & 0.288 & 0.440
& \underline{23.20} & \underline{0.715} & 0.325 \\
& ObjectGS~\cite{objectgs}
& \underline{0.178} & \underline{0.337} & \underline{0.489}
& \textbf{24.34} & \textbf{0.733} & \textbf{0.310} \\
\midrule
\textit{Feed-forward + optimization} &
IGGT~\cite{iggt} + LUDVIG~\cite{ludvig}
& 0.122 & 0.265 & 0.442
& 22.75 & 0.712 & \underline{0.323} \\
\midrule
\textit{Feed-forward} &
Ours
& \textbf{0.235} & \textbf{0.438} & \textbf{0.564}
& 22.41 & 0.709 & 0.355 \\
\bottomrule
\end{tabular}





\vspace{-10pt}
\end{table*}

\subsection{Applications: entity-level interfaces and operations}
\begin{figure}[t]
\centering
\includegraphics[width=1\linewidth]{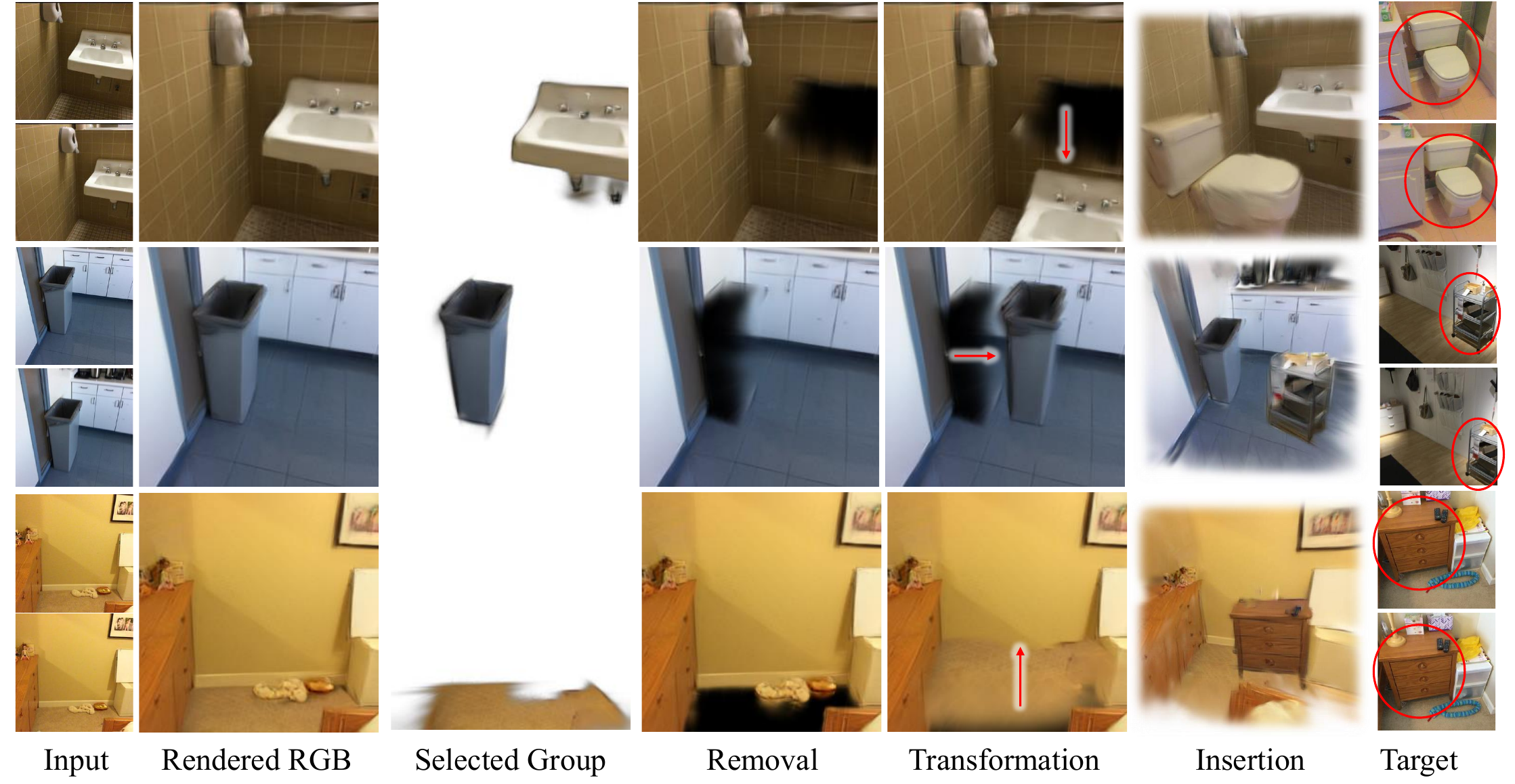}
   \vspace{-15pt}
   \caption{\textbf{Instance-level token manipulation results.} Our token groups directly offer an entity-level interface, enabling instance-level rendering, transformation, insertion, and removal.
   }
   \label{fig:token_manipulation}
   \vspace{-10pt}
\end{figure}
The token groups produced by our framework are not merely a representational convenience — they expose a direct entity-level interface through which downstream operations follow naturally, without additional modules, masks, or optimization. We demonstrate two such operations: instance-level scene manipulation and open-vocabulary 3D instance retrieval.

\textblock{Token group manipulation.}
Because each token group represents exactly one instance, scene editing reduces to selecting a group and applying an elementary operation directly to its tokens and associated Gaussians. We demonstrate four such operations: \emph{group-wise rendering} (rendering only the Gaussians of a selected group), \emph{removal} (discarding a group), \emph{insertion} (adding a group from another scene), and \emph{transformation} (applying a rigid transform to a selected group). As shown in \Fref{fig:token_manipulation}, edits stay strictly localized to the targeted instance in all cases — neighboring objects and the background are entirely unaffected. Crucially, none of these operations require manually provided masks, post-hoc processing, or any per-scene optimization; they act directly on the token groups produced in a single forward pass. This stands in contrast to prior methods, where object-level editing requires either per-scene optimization with identity supervision~\cite{gaussiangrouping, objectgs} or explicit lifting and merging of 2D predictions into 3D~\cite{iggt}. We note that the demonstrated scenes contain relatively well-separated objects; how manipulation holds up under heavy occlusion or object contact is an interesting direction for future work.

\begin{figure}[]
\centering
\includegraphics[width=1\linewidth]{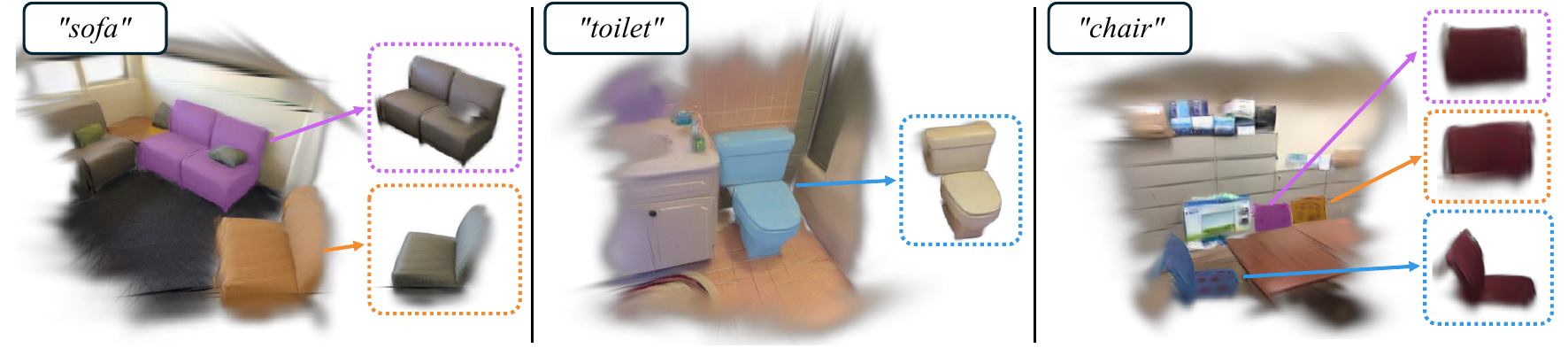}
    \vspace{-13pt}
   \caption{
   \textbf{Open-vocabulary 3D instance retrieval.} We present our results with lifted LSeg features. Our token groups naturally offer efficient instance-level retrieval operations without post-processing.
   }
   \label{fig:retrieval}
   \vspace{-5pt}
\end{figure}

\textblock{Open-vocabulary 3D instance retrieval.}
Each token group stores a shared group-level semantic embedding lifted from a 2D foundation model, enabling retrieval by matching a text or image query directly against the group embeddings. Because retrieval operates at the instance level rather than the primitive level, complexity scales linearly with the number of instances --- fewer than 100 in our representation --- rather than with the number of Gaussians, which reaches 131,072 in pixel-aligned baselines. As shown in \Fref{fig:retrieval}, a query retrieves complete, spatially coherent instances rather than a scattered subset of primitives, a result that follows directly from the instance structure of the representation rather than from any post-hoc aggregation. While the demonstrations here use unambiguous queries (\eg, sofa, toilet), the underlying mechanism generalizes to any feature expressible by the foundation model, including finer-grained or relational queries.

\subsection{Ablations}
To examine the effects of our central design choices, we conduct ablation studies on joint training and decomposed feature lifting. All ablations are conducted on ScanNet with 2 context views.

\textblock{Joint training.}
We evaluate our joint training scheme against two variants: (1) a \emph{sequential} variant that first trains $\mathcal{D}_{\mathrm{anchor}}$ with the rendering loss, then freezes it and trains $\mathcal{D}_{\mathrm{group}}$ with the mask loss; and (2) a joint variant \emph{without $\lambda_{\mathrm{seg}}$ warm-up}. As reported in \Tref{tab:ablation_joint}, the sequential variant shows substantially degraded performance on both reconstruction and segmentation — segmentation AP drops from 0.193 to 0.032 — indicating that the two decoders must be trained together for token group structure to emerge properly. The interaction between reconstruction and grouping supervision is not one-directional: joint training shapes the anchor tokens toward geometry that is compatible with coherent grouping, rather than optimizing reconstruction in isolation. Perhaps more surprisingly, removing the warm-up hurts performance even further than sequential training. This is because applying full segmentation supervision before the reconstruction branch has converged introduces conflicting gradients early in training, destabilizing both objectives simultaneously. The warm-up resolves this by allowing initial geometry to emerge before grouping supervision takes full effect, after which both objectives reinforce rather than compete with each other.

\textblock{Decomposed feature lifting.}
\Tref{tab:ablation_decomp} compares three variants: anchor residuals only (Anchor), group-level features only (Group), and our full decomposition (Group + Anchor). The anchor-only variant performs worst, as low-dimensional residuals lack the capacity to represent full semantic content without the shared group embedding to anchor them. The group-only variant already achieves strong results, which is itself an informative finding: it implies that within-group semantic variation is relatively small, validating our design choice of a compact shared embedding as the primary semantic carrier. The full model gains a further improvement by adding anchor residuals to capture the sub-instance variation that the shared embedding cannot represent. Together these results confirm the division of labor built into our representation — group features handle instance-level semantics, anchor residuals handle local specificity — and show that both levels contribute meaningfully. We note that all ablations are conducted under the 2-view setup; we expect the conclusions to hold for the 8-view setting given the consistent trends across configurations.

\begin{table*}[t]
\centering
\begin{minipage}[t]{0.62\textwidth}
\centering
\footnotesize
\setlength{\tabcolsep}{4pt}
\caption{
\textbf{Ablations on the joint reconstruction-segmentation scheme.} `Sequential' denotes sequentially training reconstruction and then segmentation. `w/o warm-up' denotes the variant without $\lambda_{\mathrm{seg}}$ warm-up in early training.
We report results on target viewpoints with 2 context views.
}
\label{tab:ablation_joint}
\begin{tabular}{l ccc ccc}
\toprule
 & \multicolumn{3}{c}{Reconstruction} & \multicolumn{3}{c}{Instance Segmentation} \\
Training & PSNR$\uparrow$ & SSIM$\uparrow$ & LPIPS$\downarrow$ & AP$\uparrow$ & AP$_{50}\uparrow$ & AP$_{25}\uparrow$ \\
\midrule
Sequential & 23.65 & 0.737 & 0.348 & 0.032 & 0.097 & 0.315 \\
w/o warm-up & 23.09  & 0.732 & 0.329 & 0.081 & 0.186  & 0.415  \\
Ours & \textbf{25.11} & \textbf{0.769} & \textbf{0.240} & \textbf{0.193} & \textbf{0.377} & \textbf{0.529} \\
\bottomrule
\end{tabular}
\end{minipage}
\hfill
\begin{minipage}[t]{0.36\textwidth}
\centering
\footnotesize
\setlength{\tabcolsep}{7pt}
\caption{
\textbf{Ablations on the decomposed feature lifting.}
`Anchor' denotes low-dimensional anchor residuals, and `Group' denotes the group-level shared feature.
We report feature lifting results on target views.
}
\label{tab:ablation_decomp}
\begin{tabular}{l cc}
\toprule
Feat. & mIoU$\uparrow$ & Acc$\uparrow$ \\
\midrule
Anchor & 0.524 & 0.713 \\
Group & 0.635 & 0.767 \\
Group + Anchor &  \textbf{0.657} & \textbf{0.789} \\
\bottomrule
\end{tabular}
\end{minipage}
\vspace{-8pt}
\end{table*}

\section{Discussion}
This work suggests a new direction for 3D scene representation: rather than reconstructing dense primitives and recovering structure post-hoc, we treat object instances as a native interface.
Below we discuss future extensions that this token group could enable, along with its current limitations.

\textblock{Toward compositional reasoning and generation.}
Since our representation encodes a scene as a compact set of object-aligned token groups, it offers a promising starting point for connecting 3D scenes to large models. For reasoning, a large language model could treat a scene as a small set of entities, operating on group-level tokens and drawing on anchor tokens when finer detail is needed. For generation, since groups are organized around instances rather than spatial primitives, a generative model could synthesize each group independently, compose multiple groups into a scene, and transfer or mix groups across scenes. These directions suggest the token group as an entity-level 3D representation for compositional reasoning and generation, bridging 3D scenes and large models.

\textblock{Toward object-centric world models for robotics.}
A robotic agent interacting with the physical world must reason about objects --- which ones are present, where they are, and how they can be manipulated --- yet dominant scene representations expose no such structure natively. Our token groups offer a natural bridge: given a handful of unposed images, the framework produces a compact set of instance-level handles that map directly onto the entities a robot needs to reason about. For perception, the group-level semantic embeddings support grounding natural language instructions ("pick up the chair") to specific token groups without additional modules. For planning, the instance-level manipulation interface --- removing, inserting, and transforming groups --- provides exactly the kind of object-level mental simulation a robotic world model needs to evaluate candidate actions before executing them. For efficiency, operating over fewer than 100 instance tokens rather than tens of thousands of primitives makes forward prediction far more tractable at the timescales robotics demands. Together these properties suggest that instance-structured token groups could serve as the perceptual front-end of an object-centric world model, connecting raw multi-view observations to the entity-level representations that planning and manipulation policies can most naturally operate over. 
Extending the framework to dynamic scenes with moving objects and real-time inference would be a necessary step toward this vision.

\textblock{Limitations.}
Our evaluation focuses primarily on bounded indoor scenes, and scaling to outdoor environments and large-scale scenes remains an open challenge --- the fixed upper bound of $L=100$ groups and the model training both likely to require revisiting at larger scale. The single shared group-level token may also have insufficient expressivity to fully capture the semantics of complex or highly varied instances; employing multiple shared semantic tokens as basis features within each group is a natural extension. Finally, the current framework assumes static scenes; extending to dynamic settings with moving objects is a prerequisite for the robotic applications discussed above.
\section{Conclusion}
We presented a feed-forward framework that reconstructs 3D scenes as instance-structured token groups, making object instances a first-class element of the representation. Within each group, an instance token summarizes entity-level identity while anchor tokens encode local geometry and appearance. The resulting token groups support accurate reconstruction and scene understanding, while directly enabling instance-level manipulations and retrieval without post-hoc processing.
\clearpage

{\small
\bibliographystyle{plain}
\bibliography{egbib}
}
\newpage
\appendix
\section{Societal impacts}
Instance-structured 3D representations benefit robotics, AR/VR, and content creation by making scenes editable at the object level. However, reconstructing and recomposing real environments from casual captures raises privacy concerns, and token-level editing lowers the barrier to fabricated 3D scenes. We encourage provenance signals, consented capture, and disclosure of edits, and view detection of manipulated 3D content as an important complement to the capabilities introduced here.

\subsection{Additional implementation details}
All feature dimensions are set to 1024, and our model takes $256\times256$ images as input and renders at the same resolution.
The Gaussian head predicts, for each anchor, the scale, rotation, opacity, and per-Gaussian local offsets relative to the anchor center, together with spherical harmonics coefficients of degree~2.
For the VGGT backbone, we follow Uni3R and change the initial DINO patch size from 14 to 16 to accommodate $256\times256$ inputs, and add a linear layer that takes the ground-truth camera intrinsics as an additional input.
We implement our framework in PyTorch with \texttt{bf16} mixed precision and rasterize the Gaussians with \texttt{gsplat}~\cite{ye2025gsplat}.
All models are optimized with AdamW (learning rate $1\times10^{-4}$, weight decay 0.05) using a linear warm-up followed by cosine decay to $1\times10^{-6}$, with gradient clipping at 0.5.
For semantic feature lifting, we train an additional group-token decoder of 4 cross-attention transformer layers on top of the frozen tokenizer.
Training the feature-lifting model takes less than 3 hours on 4 RTX A6000 GPUs under the 2-view setup.

\section{Experiment setup details}
\label{sec:setup_details}

\subsection{ScanNet class-agnostic novel-view instance segmentation}
\paragraph{Common setup.}
We follow the 40-scene ScanNet test subset introduced by LSM~\cite{lsm}. For each scene, we sample frames from the full sequence with stride 10 and reconstruct scene-level COLMAP~\cite{colmap} cameras from the sampled frames. Among the sampled frames, we use an interleaved split with 8 training views and 7 test views. All compared methods share the same COLMAP initialization and the same train/test view split, and evaluation is conducted on the test views.

\paragraph{Per-scene Gaussian labeling baselines.}
Gaussian Grouping~\cite{gaussiangrouping} and ObjectGS~\cite{objectgs} are trained in a per-scene manner using the shared COLMAP initialization described above. Their predictions are rendered on the same test views for evaluation. For both Gaussian Grouping and ObjectGS, AP confidence is computed from rendered per-pixel scores.

\paragraph{IGGT + LUDVIG with mask-regularized 3DGS.}
For IGGT~\cite{iggt} + LUDVIG~\cite{ludvig}, we use a mask-regularized 3DGS backbone instead of vanilla 3DGS for label uplifting. This follows the panoptic regularization strategy adopted in PanSt3R~\cite{panst3r}, which helps reduce the tendency of Gaussians optimized only with RGB reconstruction to spread across object boundaries. We use the IGGT instance masks as input labels and convert them into one-hot instance labels. We then uplift these labels to the mask-regularized 3DGS scene using LUDVIG. The uplifted labels are projected to the test views to obtain the final novel-view instance predictions. For this method, the confidence used for AP computation is derived from the raw scores produced during lifting and reprojection.

\section{Additional generalization experiments}
\label{sec:additional}
We complement our main ScanNet experiments with two studies that test how our tokenization framework behaves beyond the in-domain setting: (i) training on RealEstate10K with SAM2 pseudo-labels in place of human-annotated masks, and (ii) zero-shot transfer of a ScanNet-trained model to MipNeRF360. Both experiments use the 2-view input setting.

\subsection{RealEstate10K with SAM2 pseudo-labels}
\paragraph{Dataset and view sampling.}
RealEstate10K~\cite{re10k} consists of casually captured real-estate video clips with camera poses but no instance annotations. We follow the standard train/test scene split and use a 2-view setup throughout: for each clip, we sample two source views as input and evaluate reconstruction on held-out target views from the same clip. 

\paragraph{SAM2 pseudo-label generation.}
Since RE10K provides no ground-truth instance masks, we generate pseudo-labels with SAM2~\cite{sam2}. 
The resulting masks are used as the targets for our 2D instance segmentation loss; no human-annotated 3D or 2D labels are used at any stage of training. While SAM2 maintains object identities across frames within a clip, ID tracks are occasionally broken---e.g., when an object is briefly occluded or leaves and re-enters the view, it may be assigned a new ID upon reappearance, splitting a single physical instance across multiple supervisory IDs.

\paragraph{Training and baselines.}
We train our tokenizer on RE10K from scratch under the 2-view setup, keeping the loss formulation and hyperparameters identical to the ScanNet experiments unless noted otherwise. 
For reconstruction comparison, we evaluate C3G~\cite{c3g} on the same 2-view splits.

\paragraph{Results.}
\begin{table}[t]
\centering
\small
\setlength{\tabcolsep}{8pt}
\caption{Reconstruction quality on RealEstate10K (2-view). Our model is trained with SAM2 pseudo-labels as instance supervision, while baselines follow their original training protocols. \textbf{Bold} indicates the best result.}
\label{tab:re10k_recon}
\vspace{4pt}
\begin{tabular}{l ccc}
\toprule
Method & PSNR$\uparrow$ & SSIM$\uparrow$ & LPIPS$\downarrow$ \\
\midrule
C3G~\cite{c3g}     & 22.39 & 0.713          & 0.259 \\
Ours               & \textbf{22.85} & \textbf{0.746} & \textbf{0.230} \\
\bottomrule
\end{tabular}
\end{table}
\begin{figure}[]
\centering
\includegraphics[width=1\linewidth]{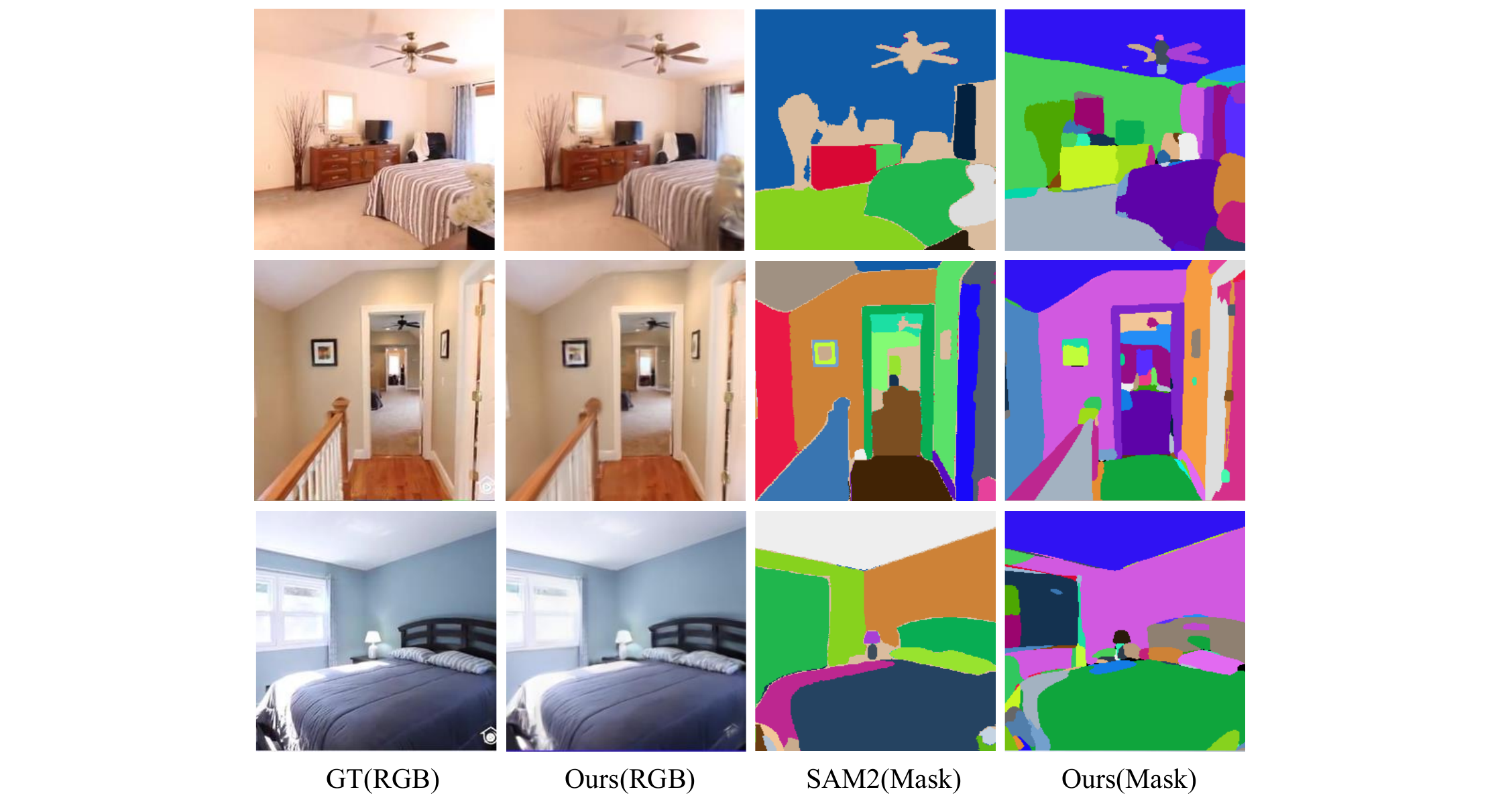}
    \vspace{-12pt}
   \caption{Qualitative results on RealEstate10K. From left to right: ground-truth RGB, our rendered RGB, our predicted instance masks, and---for reference---SAM2 masks obtained on the test view under the same protocol used to generate our training supervision. The SAM2 column is shown only to give a sense of the kind of supervisory signal our model was trained from, and is not part of the evaluation.}
   \label{fig:re10k}
   \vspace{-10pt}
\end{figure}
Table~\ref{tab:re10k_recon} reports reconstruction performance on RealEstate10K under the 2-view setting. Despite relying solely on SAM2-generated pseudo-labels for instance supervision, our tokenizer outperforms C3G across all three metrics, indicating that the anchor--group tokenization remains effective even when supervised with noisy, automatically generated masks. Figure~\ref{fig:re10k} shows our rendered RGB and predicted instance masks on held-out test views, along with SAM2 masks on the same views shown for reference---to give a sense of the kind of supervisory signal the model was trained from, not as a direct baseline. Our token groups produce coherent instance decompositions on scenes that lack ground-truth annotations altogether, and the decompositions remain consistent even where the reference SAM2 masks fragment a single physical instance into multiple IDs. Together, these results indicate that the instance structure emerging in our representation does not depend on dataset-specific clean labels, and can be bootstrapped from off-the-shelf 2D segmentation models.

\subsection{Zero-shot transfer to MipNeRF360}
\label{sec:appendix:mipnerf360}
\paragraph{Setup.}
We evaluate zero-shot generalization to MipNeRF360~\cite{mipnerf360}, applying our ScanNet-trained model directly without any fine-tuning. We use a 2-context, 1-target setup: for each scene, two source views are provided as input and reconstruction is evaluated on a single held-out target view. 

\paragraph{Baseline.}
We compare against Uni3R~\cite{uni3r} under the same zero-shot protocol, also applying its ScanNet-trained checkpoint directly to MipNeRF360 without fine-tuning. Both methods share the same 2-context, 1-target splits and evaluation views. 

\paragraph{Results.}
\begin{table}[t]
\centering
\small
\setlength{\tabcolsep}{8pt}
\caption{Zero-shot reconstruction on MipNeRF360 (2 context views, 1 target view). All models are trained on ScanNet and evaluated on MipNeRF360 without any fine-tuning. \textbf{Bold} indicates the best result.}
\label{tab:mipnerf360_recon}
\vspace{4pt}
\begin{tabular}{l ccc}
\toprule
Method & PSNR$\uparrow$ & SSIM$\uparrow$ & LPIPS$\downarrow$ \\
\midrule
Uni3R~\cite{uni3r} & 14.58 & 0.317 & 0.472 \\
Ours               & \textbf{16.52} & \textbf{0.408} & \textbf{0.439} \\
\bottomrule
\end{tabular}
\end{table}
Table~\ref{tab:mipnerf360_recon} reports PSNR, SSIM, and LPIPS on the target views. Our model outperforms Uni3R across all three metrics despite the marked distribution shift from ScanNet's bounded indoor layouts to MipNeRF360's unbounded outdoor and object-centric scenes, indicating that the anchor--group tokenization captures a representation prior that holds beyond the training distribution.

\section{Additional qualitative results}
\label{sec:additional_qual}
We provide additional qualitative comparisons on three more ScanNet scenes each for novel-view reconstruction, open-vocabulary feature lifting, and class-agnostic instance segmentation, complementing the main-paper figures.

\subsection{Additional reconstruction results}
\begin{figure}[t]
\centering
\includegraphics[width=1\linewidth]{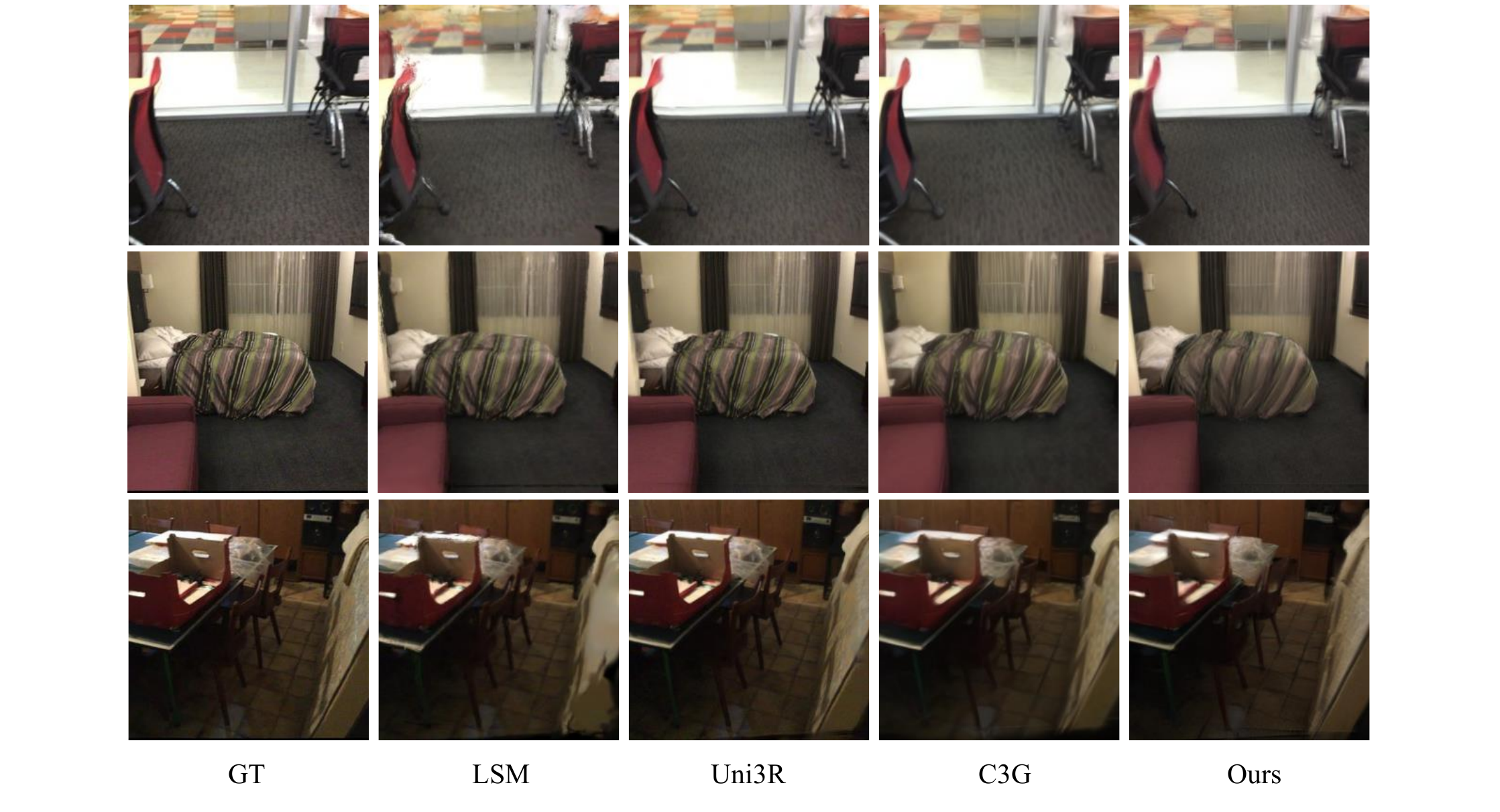}
\vspace{-12pt}
\caption{Additional qualitative reconstruction results on ScanNet, complementing Figure~\ref{fig:recon} of the main paper. Three additional scenes are shown; each row presents a single test view rendered by LSM~\cite{lsm}, Uni3R~\cite{uni3r}, C3G~\cite{c3g}, and ours, alongside the ground truth.}
\label{fig:recon_supp}
\vspace{-10pt}
\end{figure}
Figure~\ref{fig:recon_supp} provides additional novel-view reconstruction comparisons on three more ScanNet scenes, extending the qualitative comparison in Figure~\ref{fig:recon} of the main paper. Across diverse layouts, our anchor--group tokenization reproduces the overall scene structure and major object appearances faithfully, while the compact token-based representation occasionally smooths over fine high-frequency details compared to pixel-aligned baselines---consistent with the small reconstruction gap reported in Table~\ref{tab:recon_feature}.

\subsection{Additional feature lifting results}
\begin{figure}[t]
\centering
\includegraphics[width=1\linewidth]{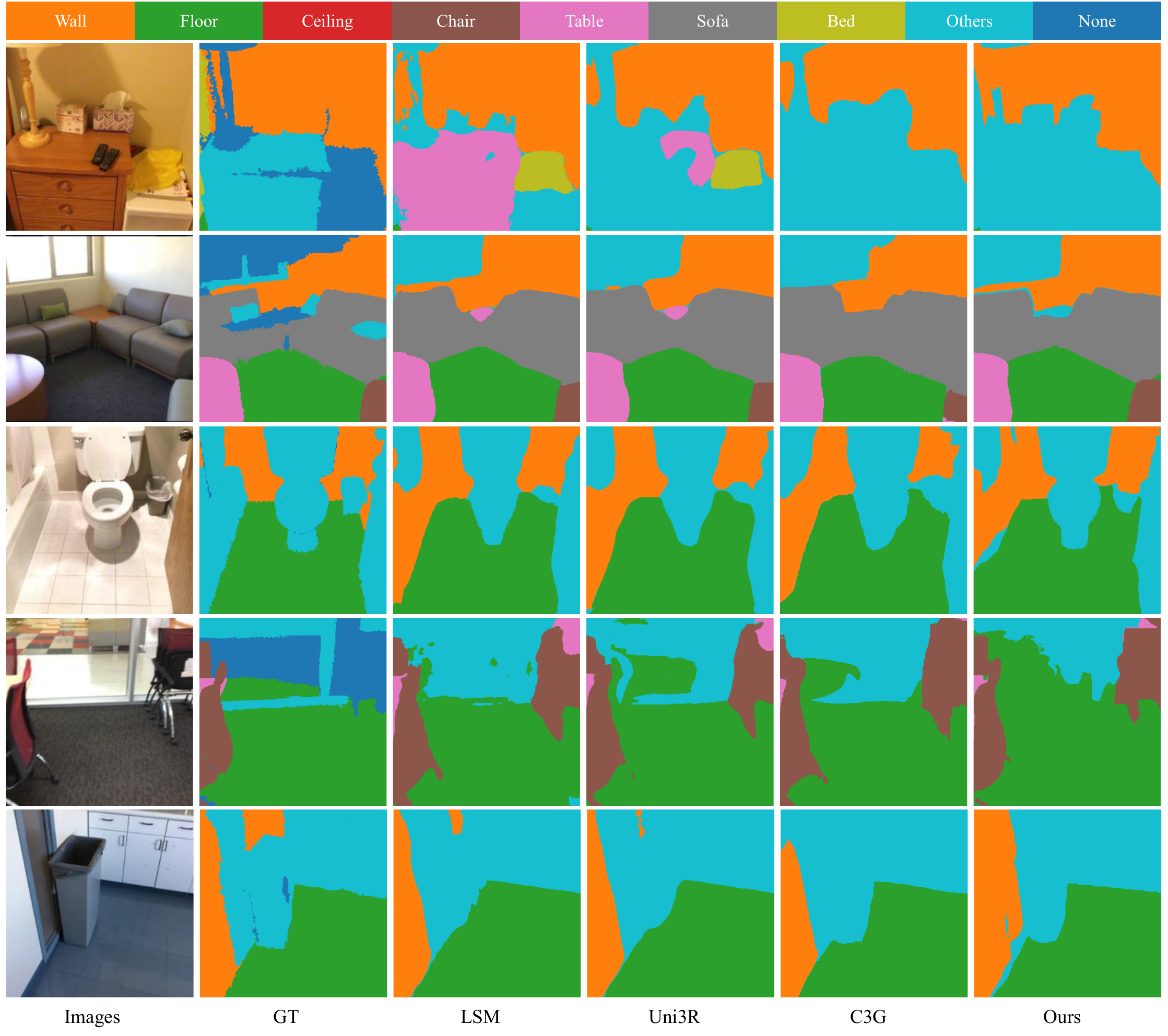}
\vspace{-5pt}
\caption{Additional qualitative LSeg~\cite{lseg} feature distillation results on ScanNet, complementing Figure~\ref{fig:sem} of the main paper. Three additional scenes are shown; each row presents the lifted semantic feature map from LSM~\cite{lsm}, Uni3R~\cite{uni3r}, C3G~\cite{c3g}, and ours, alongside the ground-truth segmentation.}
\label{fig:sem_supp}
\vspace{-5pt}
\end{figure}
Figure~\ref{fig:sem_supp} provides additional open-vocabulary feature lifting comparisons on three more ScanNet scenes, extending the qualitative comparison in Figure~\ref{fig:sem} of the main paper. The decomposed group-level and anchor-level features predicted by our model produce semantic maps that respect object boundaries and remain consistent across regions belonging to the same entity, supporting the quantitative gains in source- and target-view mIoU reported in Table~\ref{tab:recon_feature}.

\subsection{Additional class-agnostic instance segmentation results}
\begin{figure}[t]
\centering
\includegraphics[width=1\linewidth]{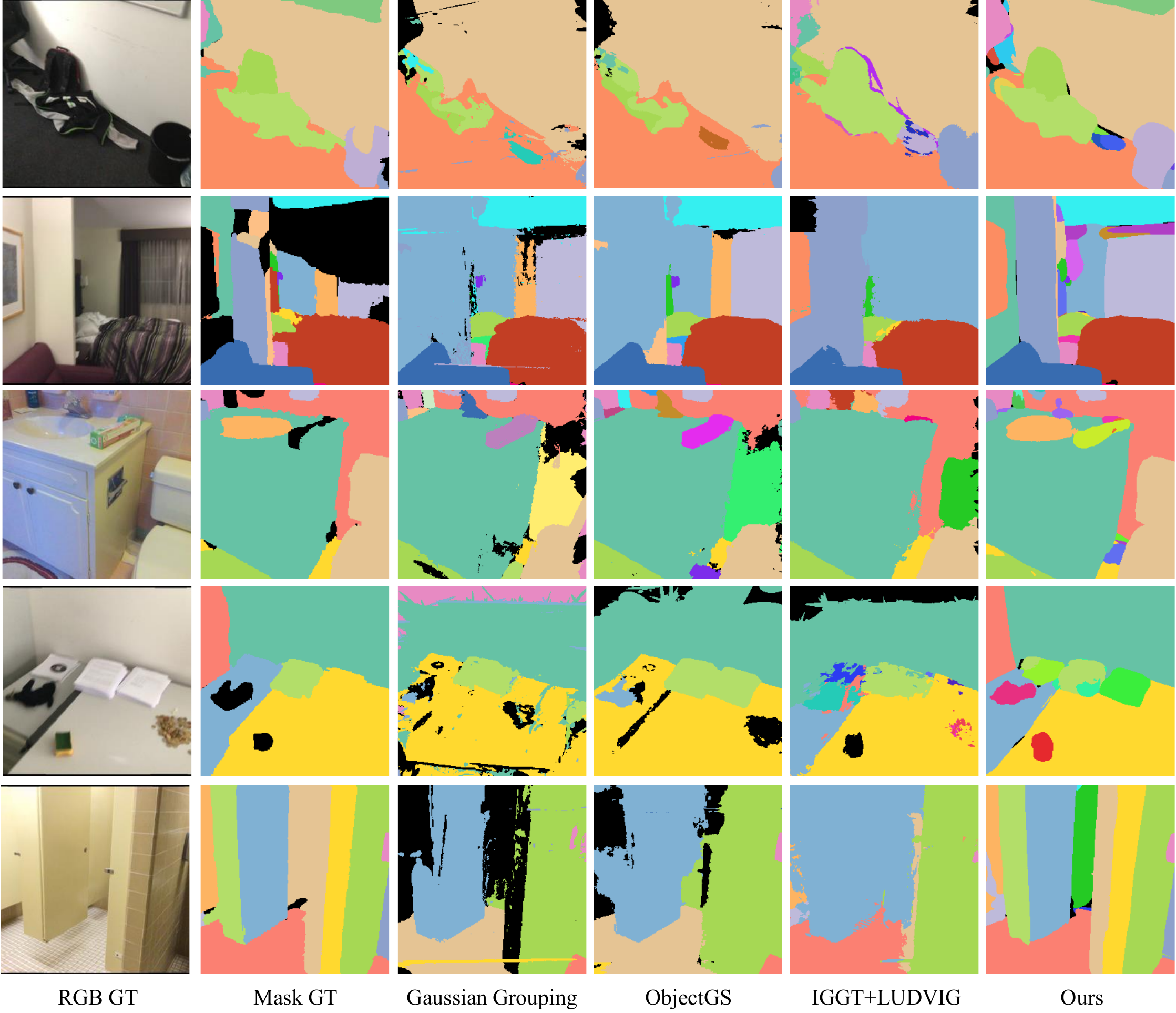}
\vspace{-12pt}
\caption{Additional qualitative class-agnostic novel-view instance segmentation results on ScanNet, complementing Figure~\ref{fig:class_agnostic_seg} of the main paper. Three additional scenes are shown; each row presents the predicted instance masks from Gaussian Grouping~\cite{gaussiangrouping}, ObjectGS~\cite{objectgs}, IGGT~\cite{iggt}+LUDVIG~\cite{ludvig}, and ours, alongside the ground-truth RGB and instance mask.}
\label{fig:class_agnostic_seg_supp}
\vspace{-10pt}
\end{figure}
Figure~\ref{fig:class_agnostic_seg_supp} provides additional class-agnostic instance segmentation comparisons on three more ScanNet scenes, extending the qualitative comparison in Figure~\ref{fig:class_agnostic_seg} of the main paper. Our token-group-based segmentation continues to produce coherent instance boundaries and avoids the fragmented regions characteristic of per-Gaussian identity baselines, particularly on large, contiguous surfaces such as walls, floors, and beds.

\end{document}